\documentclass[%
 reprint,
 amsmath,amssymb,
 aps,
]{revtex4-1}

\usepackage{graphicx}
\usepackage{dcolumn}
\usepackage{bm}

\begin{document}

\preprint{APS/123-QED}

\title{Fractal dimension analysis for automatic morphological galaxy classification}

\author{Jorge de la Calleja}
\affiliation{%
 Universidad Polit\'ecnica de Puebla, 72640, M\'exico\\
}%


\author{Elsa M. de la Calleja}
\affiliation{Instituto de Investigaciones en Materiales, Universidad Nacional Aut\'onoma de M\'exico, Apdo. Postal 70-360, Ciudad Universitaria, 04510, M\'exico.\\
}%
\author{Hugo Jair Escalante}
\affiliation{%
 Instituto Nacional de Astrof\'isica, \'Optica, Electr\'onica, Computer Science Department, Puebla, M\'exico, 72840\\
}%



\begin{abstract}
In this report we present experimental results using \emph{Haussdorf-Besicovich} fractal dimension for performing morphological galaxy classification. The fractal dimension is a topological, structural and spatial property that give us information about the space were an object lives. We have calculated the fractal dimension value of the main types of galaxies: ellipticals, spirals and irregulars; and we use it as a feature for classifying them. Also, we have performed an image analysis process in order to standardize the galaxy images, and we have used principal component analysis to obtain the main attributes in the images. Galaxy classification was performed using machine learning algorithms: C4.5, k-nearest neighbors, random forest and support vector machines. Preliminary experimental results using 10-fold cross-validation show that fractal dimension helps to improve classification, with over 88 per cent accuracy for elliptical galaxies, 100 per cent accuracy for spiral galaxies and over 40 per cent for irregular galaxies. 
\end{abstract}

\pacs{galaxies, fractal dimension, machine learning}
\maketitle


\section{Introduction}
Astronomy has a long history of acquiring and analyzing enormous quantities of data. As many other fields, this science has become very data-rich due to advances in telescope, detector, and computer technology. Recently, numerous digital sky surveys across a wide range of wavelengths are producing very large image databases of astronomical objects. For example, the Large Synoptic Sky Survey will produce billions of galaxy images \cite{Abell:2009jd}. Therefore there is a need to create robust and automated tools for processing astronomical data, particularly for the analysis of the morphology of celestial objects such as galaxies.

Edwin Hubble in 1926 devised a formal galaxy classification approach, known as the Hubble tuning-fork. In his classification scheme galaxies are classified based on their shape, and there are three basic types: elliptical galaxies, spiral galaxies, and irregular galaxies. The morphology of galaxies is generally an important issue in the large scale study of the Universe. Galaxy classification is the first step towards a greater understanding of the origin and formation process of galaxies, and also the evolution process of the Universe. Visual inspection for classifying galaxies has been done by experts, however, it is not easy, because it requires skill and experience, and it is also time-consuming. On the other hand, automatic classification can analyze thousands of images in seconds, and also they are more impartial than humans, i.e. they are not subject to the conscious and unconscious prejudices which affect humans in looking at galaxy images \cite{Ball:2004jd}.

Several approaches have been carried out for automatic image analysis and galaxy classification using machine learning and computer vision techniques. First approaches have used artificial neural networks \cite{Ball:2004jd,delaCalleja:2004jd,Goderya:2002jd,Sodre:1992jd}, decision trees \cite{Marin:2013jd,Owens:1996jd}, instance-based methods \cite{Shamir:2009jd}, kernel methods \cite{Freed:2015jd}, among others. Recently, some significant works have been presented using new approaches. For example, the Sparse Representation technique and dictionary learning \cite{diazHernandez:2016jd}, and deep neural networks \cite{Dieleman:2015jd}.

Classification of galaxies can help to identify with a little more exactitude the localization of some kind of galaxies and to justify their study in depth. To add elements to describe the universe objects respond to different interests; i.e., the description of the existence of geometrical nature of space-time singularities and discover the nature of the physics which takes place there~\cite{Penrose:2001jd}, the study of astrophysical jets associated with outflows originating from accretion processes in star-forming regions or galaxy clusters~\cite{Beall:2015jd} or discover new shine objects.

Generally, automated galaxy classification is performed as follows: extracting relevant information in a galaxy image, encode it as efficiently as possible, and compare one galaxy encoding with a database of similarly encoded images. This research propose the use of the fractal dimension analysis to obtain additional information in order to determine the type of galaxy. Fractal dimension is present in many objects in nature, structures generated by mathematical algorithms, spatial interactions among populations, the distributions of particles in amorphous solids, and in particle configurations created by computer simulations \cite{delaCallejaElsa:2016jd}. Thus, we can measure the fractal characteristics in a wide variety of two-dimensional digital images such as galaxy images. We have also performed an image analysis stage, which we introduced in early work \cite{delaCalleja:2004jd}. In this stage we standardized galaxy images, that is, we rotated, centered and cropped the images. In addition, we use principal component analysis (PCA) to reduce data and find features that characterize them. Finally, we have used machine learning algorithms to classify the three main types of galaxies; particularly we used C4.5, k-nearest neighbors, random forest and support vector machines.  

The remainder of this paper is organized as follows. The next section provides a theoretical background on the fractal dimension. Section~\ref{sec:hubblescheme} describes the adopted Hubble galaxy classification scheme.  Section~\ref{sec:classification_process} introduces the proposed galaxy classification methodology. Section~\ref{sec:experiments} reports experimental results on real galaxy images and Section~\ref{sec:discussion} presents a discussion of results. Finally, Section~\ref{sec:conclusions} outlines conclusions and future work directions.    

\section{Fractal dimension}
The fractal dimension is a topological, structural and spatial property that give us information about the space were an object lives. The image of a galaxy is composed by many objects that can be apparently close to each other, however this is not true. The distance between them can be of billion kilometers, nevertheless, its possible to classify the galaxies by taking into account the objects projected in a two dimensional image.

The fractality is present in many physical systems, to measure the distribution of particles at mesoscopic scales~\cite{Weitz:1984jd} or on macroscopic scales such as the famous fractality of the Britain island~\cite{Mandelbrot:1967jd}. However, one of the basic discussions is the connectivity factor. According to the basic definition of a fractal, this property appears in connected systems. How we can extrapolate the connectivity in galaxies? The superposition of stars give us this connectivity that we required to obtain our structural parameter. The fractality property appears emerging in galaxies, taking in consideration the large scales and were apparently the connectivity its not inherent.

The local fractal dimension~\cite{Halsey:1986jd,Hentschel:1983jd,Ott:1993jd,Feigenbaum:1986jd} is calculated following the standard procedure \cite{delaCallejaElsa:2016jd} using the generalized box counting dimension defined as:

\begin{equation}
D_{Q}=\frac{1}{1-Q} lim_{\epsilon \rightarrow 0}\frac{lnI(Q,\epsilon)}{ln(\epsilon_{0}/\epsilon)}
\end{equation}

where $\epsilon$ is the size of the box which acquired successively smaller values of length until the minimum value of $\epsilon_{0}$ and  $I(Q,\epsilon)=\sum_{i}^{N(\epsilon)}[P_{i,Q}]^{Q}$ and $Q$ is a parameter which gives the width of the spectrum and when $Q=0$ the generalized fractal dimension represents the classic fractal dimension. We calculate the fractal dimension of each image by the multi-fractal spectrum obtained by the plugin \textit{FracLac} for ImageJ \cite{Ferreira:2013jd,chhabra:1989jd,Meneveau:1989jd}, that gives us the generalized fractal dimension using a gray scale differential option and the ode default sampling sizes.

\section{The Hubble tunning fork scheme}
\label{sec:hubblescheme}
\begin{figure}
\centering
\includegraphics[width=0.5\textwidth]{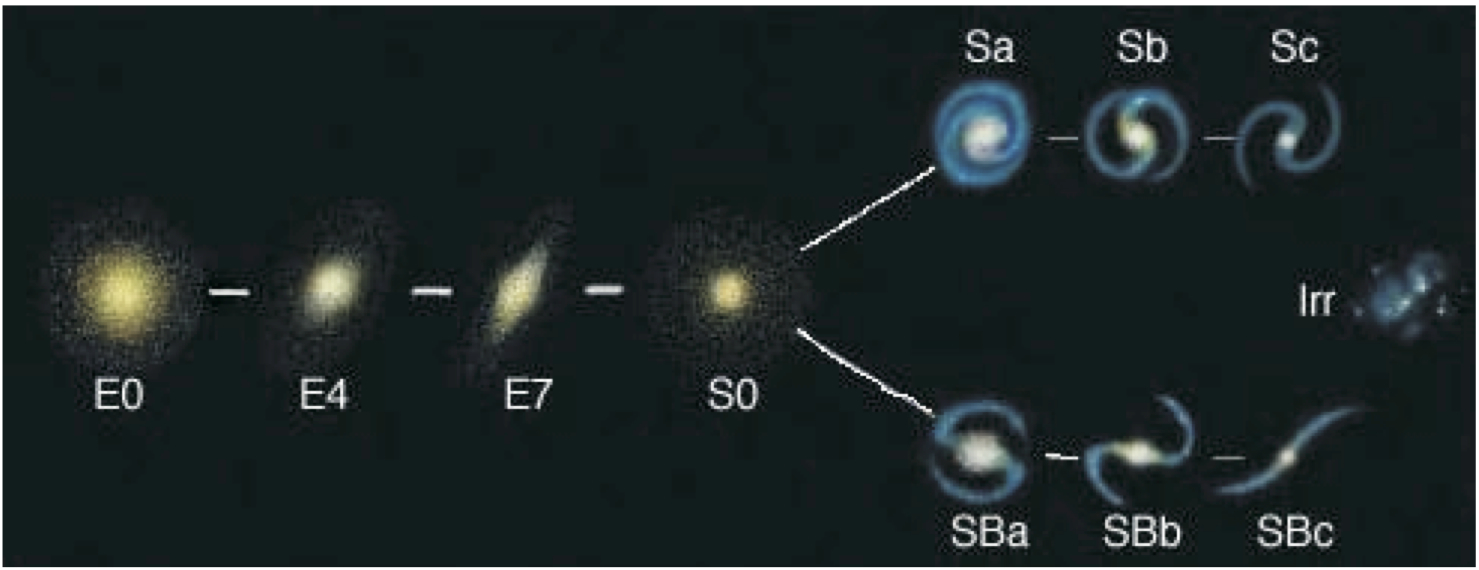}
\caption{\label{hubble-scheme} The Hubble Tuning Fork Scheme}
\end{figure}

Galaxies are large systems of stars and clouds of gas and dust, all held together by gravity. Galaxies have many different characteristics, but the easiest way to classify them is by their shape; Edwin Hubble devised a basic method for classifying them in this way \cite{Ball:2001jd}. In his classification scheme, there are three main types of galaxies: Ellipticals, Spirals and Irregulars (Figure \ref{hubble-scheme}). The different shapes of galaxies tell us something about their properties such as luminous mass, diameter, interstellar material, among others. Elliptical galaxies have the shape of an ellipsoid. Spiral galaxies are divided in ordinary and barred: ordinary spirals have an approximately spherical nucleus, while barred spirals have a elongated nucleus that looks like a bar. Finally, irregular galaxies do not have an obvious elliptical or spiral shape. In our study we have classified the main tree types, using a data set of 131 images as shown in Table~\ref{galaxyData}.

\begin{table}
\caption{Number of galaxies considered in our study.}\label{tab:data}
\label{galaxyData}
\centering
\begin{tabular}{cc}
\hline
Type         & Number of galaxies\\
\hline
Elliptical   & 17\\
Spiral       & 104\\
Irregular    & 10\\
\hline
\end{tabular}
\end{table}

\section{The classification process}
\label{sec:classification_process}
The classification process consists of three main stages: image analysis, feature extraction and automated classification. In the image analysis stage the galaxy images are rotated, centered and cropped. After that, we use principal component analysis to reduce the dimensionality of the data and to find a set of features; in addition calculating the fractal dimension of the galaxy images. Finally, these features are the input parameters for machine learning algorithms in order to classify galaxies. The next subsections describe each stage in detail.

\subsection{Image analysis}
Galaxy images generally are of different sizes and color formats, and most of the time the galaxy contained in the image is not at the center. Therefore, the aim of this stage is to create images invariant to color, position, orientation and size. We have introduced this image analysis process in early work \cite{delaCalleja:2004jd}, thus, we only present a brief description of this stage.

First, the galaxy is found in the image by applying a threshold, that is, from the original image $I$, it is generated a binary image $B$, such that we will obtain the pixels that conform the galaxy. Then we obtain $\bar{i}$ and $\bar{j}$, the center row and column of the galaxy in the image, respectively. Next we obtain the covariance matrix $C$ of the points in the galaxy image, given by

\begin {equation}\label{covMatrix}
C = \sum_{i=1}^m \sum_{j=1}^n
B(i,j)[i-\bar{i},j-\bar{j}]^T[i-\bar{i},j-\bar{j}]
\end{equation}

\begin{figure}
\centering
\includegraphics[width=0.5\textwidth]{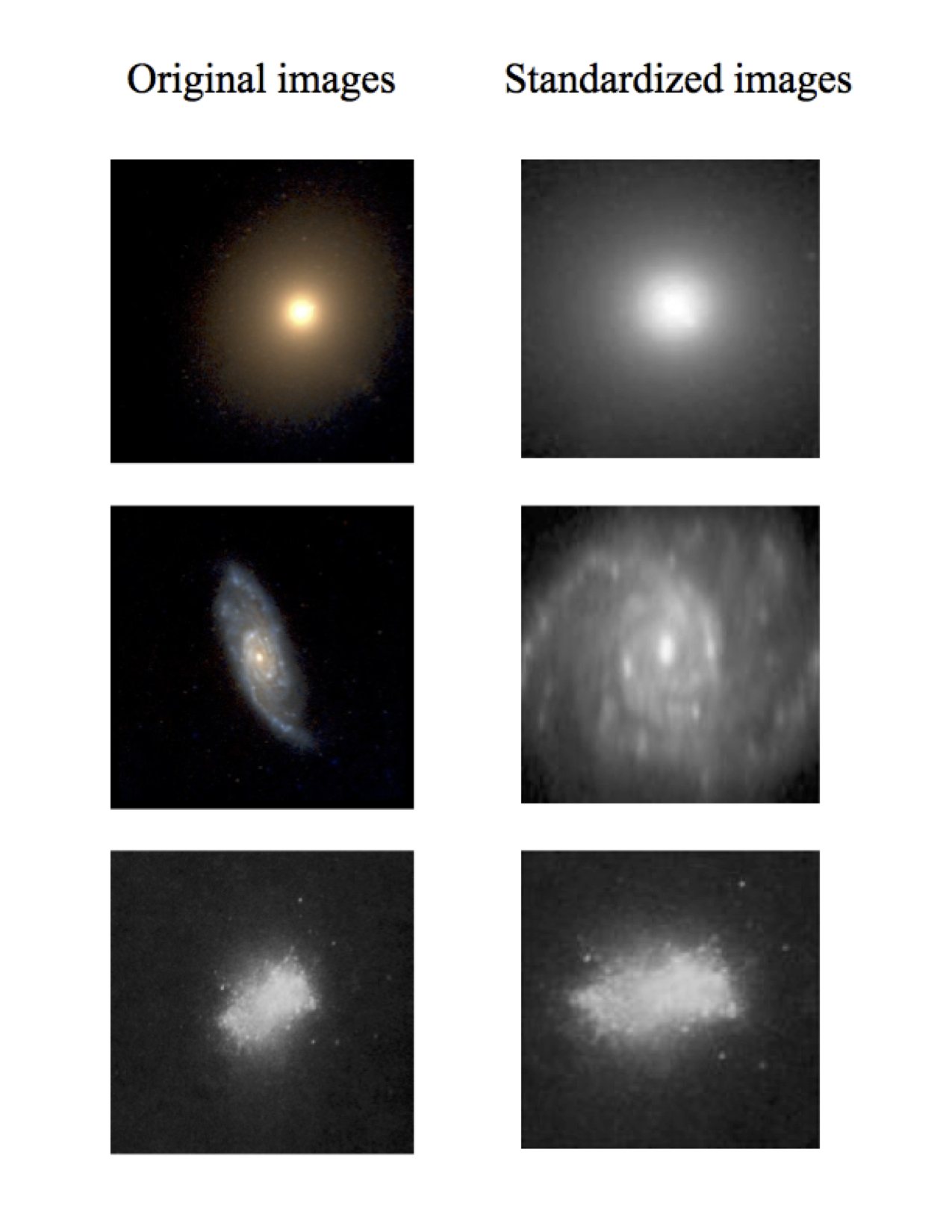}
\caption{\label{examples} Examples of Galaxies. Left: Original images, Right: Standardized images}
\end{figure}

The galaxy's main axis is given by the first eigenvector of $C$. Then the image is rotated so that the main axis is horizontal. After that the image is cropped, eliminating the columns that contain only background (black) pixels. Finally, we stretch and standardize the images to a size of 128x128 pixels. Figure \ref{examples} shows examples of original and standardized images for an elliptical galaxy, a spiral galaxy, and an irregular galaxy. 

\subsection{Feature extraction}
A general idea for galaxy classification is to extract relevant information (features) in a galaxy image, encode it as efficiently as possible, and compare one galaxy encoding with a database of similarly encoded images. However, one of the difficulties when performing this task is to find a set of characteristics that describe them as best as possible. Generally, the galaxy image is transformed into a vector with feature values. In this study, we have used principal component analysis (PCA) to find this set of relevant features, in addition we have calculated the fractal dimension of the galaxy images.

\subsubsection{Principal component analysis}
The basic idea in PCA is to find the components (the eigenvectors) of the covariance matrix of the set objects, so that they explain the maximum amount of variance possible by $n$ linearly transformed components. These eigenvectors can be thought of as a set of features which together characterize the variation among the objects \cite{TurkandPentland:1991jd}, in this case the galaxies. 

The formulation of standard PCA is as follows. Consider a set of $M$ objects $O_1,O_2,\ldots,O_M$, where the mean object of the set is defined by

\begin {equation}\label{meanEq}
\mathbf{X}= \frac{1}{M} \sum_{n=1}^M \mathbf{O}_n
\end{equation}

Each object differs from the mean by the vector

\begin {equation}\label{differenceEq}
\theta_i = O_i - X
\end{equation}

Therefore, principal component analysis seeks a set of $M$ orthogonal vectors $v$ and their associated eigenvalues $k$ which best describes the distribution of the data. The vectors $v$ and scalars $k$ are the eigenvectors and eigenvalues, respectively, of the covariance matrix

\begin {equation}\label{covMatrixEq}
C = \sum_{n=1}^M \theta_n \theta_n^T = AA^T
\end{equation}

where the matrix $A = [\theta_1,\theta_2,\ldots,\theta_M]$.

\begin{figure}
\centering
\includegraphics[width=0.5\textwidth]{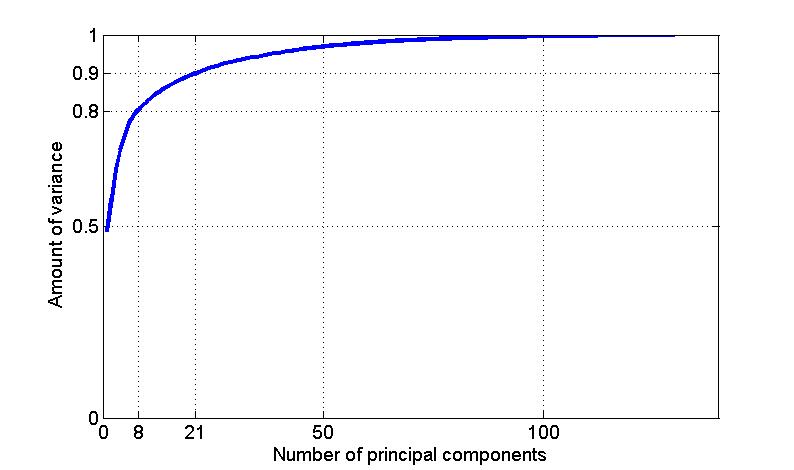}
\includegraphics[width=0.5\textwidth]{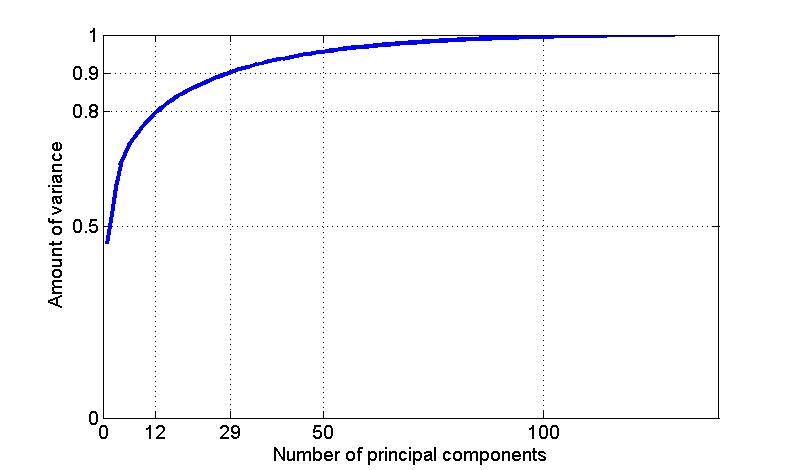}
\caption{\label{PCs} Amount of cumulative variance of the principal components. Top: Original images. Bottom: Standardized images.}
\end{figure}

The associated eigenvalues allow us to rank the eigenvectors (features) according their usefulness in characterizing the variation among the objects. In our study, we have used 8 and 12 principal components, which represent about $80\%$ of the information in original and standardized images, respectively; and 21 and 29 principal components, which represent about $90\%$ of the information in the same way (Figure \ref{PCs}). 

\subsubsection{Fractal dimension}
We obtain the generalized fractal dimension of different samples of the three main types of galaxies. In Fig.~\ref{FD}(a) is presented the fractal dimension of $17$ elliptical galaxies. Most of the values of fractal dimension are around $1.80-1.83$. The behavior of $104$ spiral galaxies looks a bit different from the previous ones which are presented in Fig.\ref{FD}(b), because most of the values fall between $1.77-1.78$. Finally in Fig.~\ref{FD}(c) the behavior of $10$ irregular galaxies obtained values around $1.774-1.776$. The ranks of fractal dimension of these three types of galaxies are evidently different. Also we can observe that the vast majority of spiral galaxies are in a very similar range; irregular galaxies also exhibit this behavior; while elliptical galaxies vary a little more.

\begin{figure}
\centering
\includegraphics[width=0.5\textwidth]{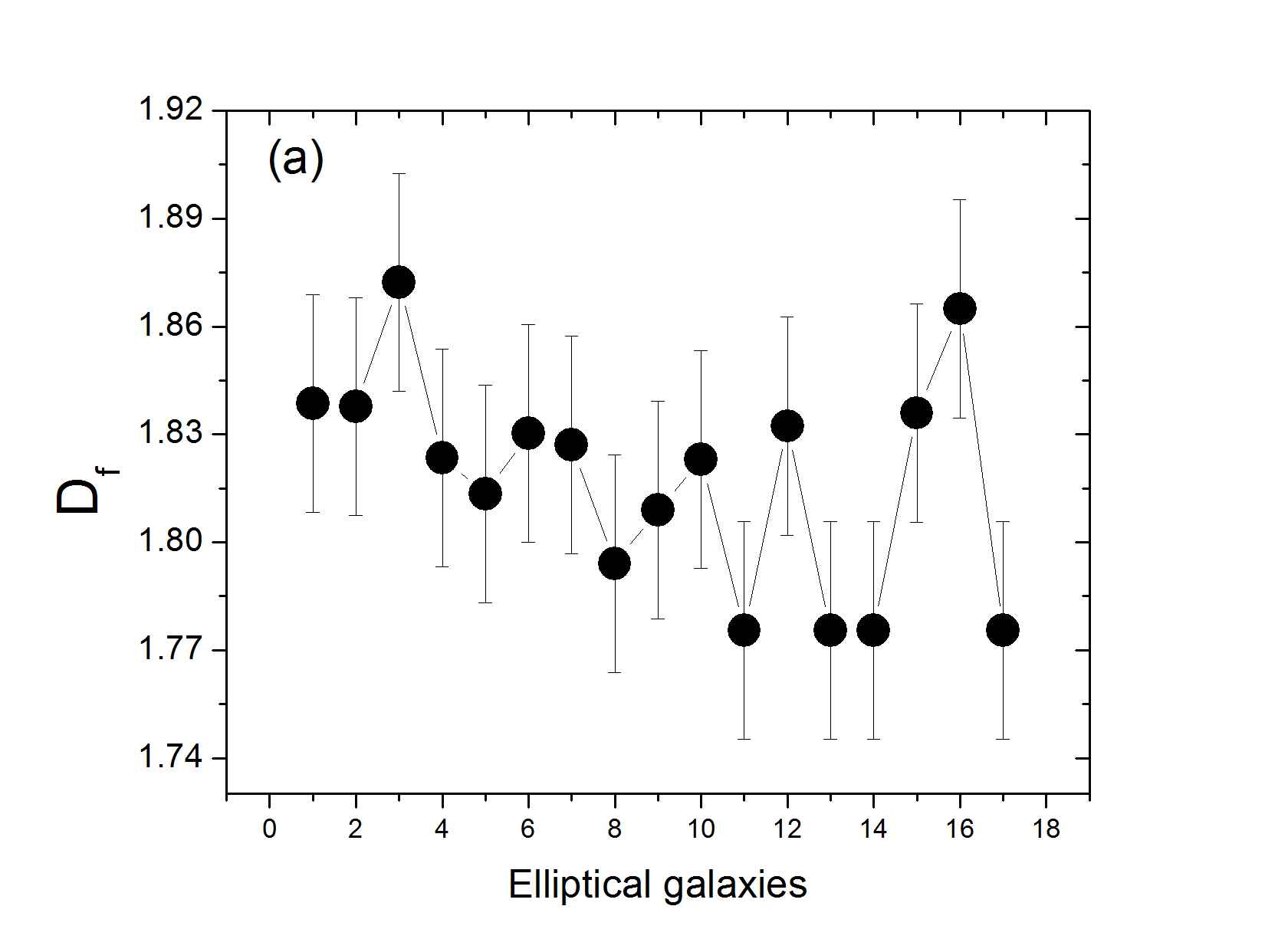}
\includegraphics[width=0.5\textwidth]{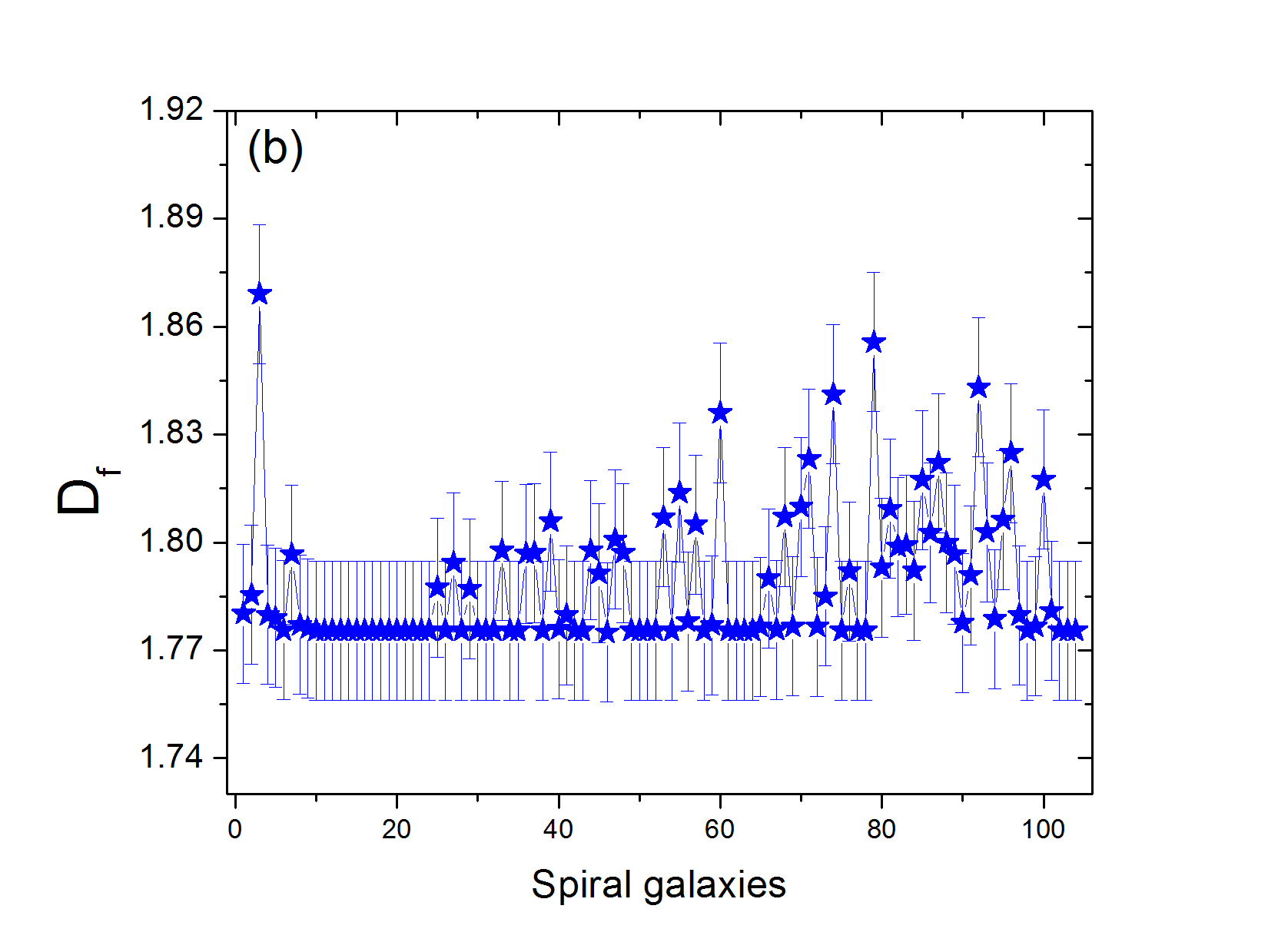}
\includegraphics[width=0.5\textwidth]{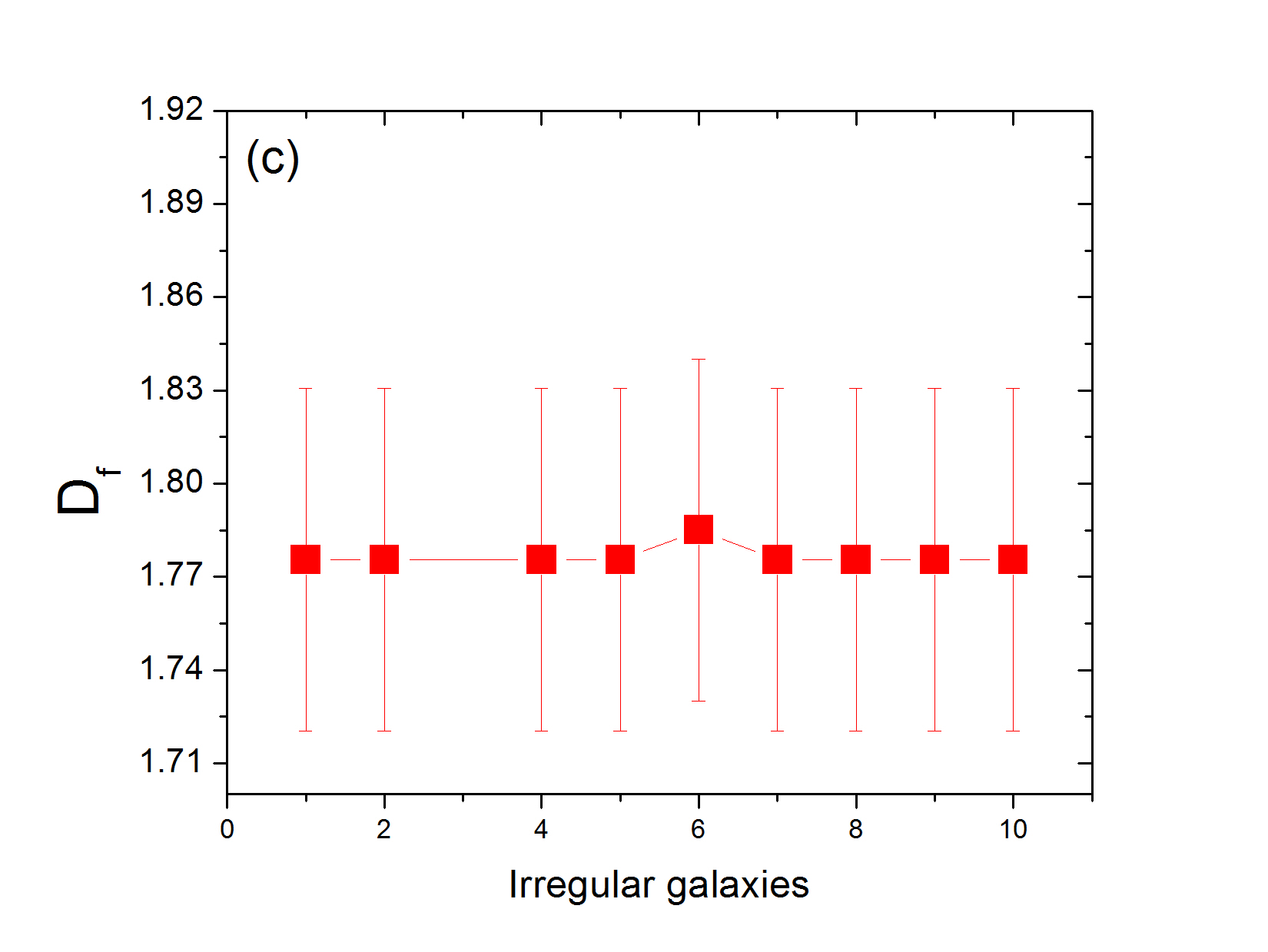}
\caption{\label{FD} In (a) black dots show the behavior of fractal dimension of elliptical galaxies. In (b) blue stars show the behavior of spiral galaxies; and finally in (c) red squares present the fractal dimension of irregular galaxies.}\label{FD}
\end{figure}

\subsection{Classification}
In order to classify the galaxy images we used the following machine learning algorithms: C4.5, k-nearest neighbors, random forest and support vector machines. Each algorithm takes as input the projection of the images onto the principal components, and also the value of the fractal dimension. Then the algorithms classify the images according to the main three types of the Hubble sequence of galaxies. Next we give a brief description of the machine learning algorithms used in this study.

\subsubsection{C4.5}
C4.5 is an extension of Quinlan's earlier ID3 algorithm used to generate a decision tree. This algorithm operates by recursively splitting a training set based on feature values to produce a tree such that each example can end up in only one leaf. An initial feature is chosen as the root of the tree, and the examples are split among branches based on the feature value for each example. If the values are continuous, then each branch takes a certain range of values. Then a new feature is chosen, and the process is repeated for the remaining examples. Then the tree is converted to an equivalent rule set, which is pruned. For a deeper introduction of this algorithm we refer the reader to \cite{Quinlan:1986jd}.

\subsubsection{k-nearest neighbors}
K-nearest neighbors (k-nn) belongs to the family of instance-based learning methods. These algorithms simply store all the available training data, and when a new query instance is encountered, they find the training examples similar to the query, and use them to classify the new query instances. K-nearest neighbors are defined in terms of standard Euclidean distance. For details of this method we refer the reader to~\cite{Mitchell:1997qr}.

\subsubsection{Random forest}
A random forest (RF) is a classifier consisting of a collection of individual tree classifiers. Each tree depends on the values of a random vector sampled independently and with the same distribution for all trees in the forest. Basically, random forest is an ensemble of unpruned trees, induced from bootstrap samples of the training data, using random feature selection in the tree induction process. Prediction is done by majority votes from predictions from the ensemble of tree. Details about this method can be found in \cite{Breiman:2001jd}.

\subsubsection{Support vector machines}
Support Vector Machines (SVMs) \cite{Vapnik:1995qr} are based on the Structural Risk Minimization principle from computational learning theory. This principle provides a formal mechanism to select a hypothesis from a hypothesis space for learning from finite training data sets. The aim of SVMs is to compute the hyperplane that best separates a set of training examples. Two cases are analyzed: the linear separable case and the non-linear separable case. In the first case we are looking for the optimal hyperplane in the set of hyper-planes separating the given training examples. The optimal hyperplane maximizes the sum of the distances to the closest positive and negative training examples (considering only two classes). The second case is solved by mapping training examples to a high-dimensional feature space using kernel functions. In this space the decision boundary is linear and we can apply the first case. There are several kernels such as polynomial, radial basis functions, neural networks, Fourier series, and splines, among others; that are chosen depending on the application.

\section{Experimental Results}
\label{sec:experiments}
The data set consisted of 131 images of galaxies, see Table~\ref{tab:data}. Most of them were taken from the NGC catalog on the web page of the Astronomical Society of the Pacific, and their classification was taken from the interactive NGC catalog on line at www.seds.org. 

The experiments were carried out using Weka, a software package that implements machine learning algorithms for data mining tasks \cite{Frank:20146qr}. In order to measure the overall accuracy of the machine learning algorithms, We used 10-fold cross-validation for all the experiments; that is, the original data set is randomly divided into ten equally sized subsets and performed 10 experiments, using in each experiment one of the subsets for testing and the other nine for training. As previously mentioned, we experiment with the following machine learning algorithms: decision trees, k-nearest neighbors,  random forest and support vector machines. For the case of decision trees, and random forest we use default parameters. For the case of k-nn we use three neighbors with weighted distance, and we use a two-degree polynomial kernel for support vector machines. 

Tables \ref{resultsOriginal} and \ref{resultsStandardized} show the accuracy for each machine learning algorithm using the original images and the standardized ones, respectively. These results were obtained by averaging the results of five runs of 10-fold cross-validation for each algorithm. As we can observe from Table \ref{resultsOriginal}, the best results were obtained by 3-nearest neighbors, with 81.82 per cent accuracy using only PCs, and 81.67 per cent accuracy using PCs plus the fractal dimension value. On the other hand, we can see from Table \ref{resultsStandardized}, that random forest obtained the best results with 85.95 and 86.71 per cent accuracy, using PCs and PCs plus the fractal dimension value, respectively.  

\begin{table}
\caption{Accuracy for original images using different number of principal components (PCs) and using the PCs plus the fractal dimension value (FDV). The best results are in bold.}
\label{resultsOriginal}
\centering
\begin{tabular}{ccccc}
\hline
  & \multicolumn{2}{c}{PCs} & \multicolumn{2}{c}{PCs + FDV}\  \\
\hline
Algorithm &      8   &   21        &   9     &   22    \\
\hline
C4.5      &   71.29  &  70.83      &  71.29  &  70.52  \\
3-nn      &   77.55  &  \bf{81.82} &  79.22  &  \bf{81.67}  \\
RF        &   78.31  &  80.91      &  80.30  &  80.60  \\
SVM       &   79.38  &  79.53      &  79.38  &  79.99  \\
\hline
mean       &   76.63  & 78.27      &  77.54  &  78.20  \\
\hline
\end{tabular}
\end{table}

\begin{table}
\caption{Accuracy for standardized images using different number of principal components (PCs) and using the PCs plus the fractal dimension value (FDV). The best results are in bold.}
\label{resultsStandardized}
\centering
\begin{tabular}{ccccc} 
\hline
  & \multicolumn{2}{c}{PCs} & \multicolumn{2}{c}{PCs + FDV}\  \\
\hline
Algorithm &     12        &   29        &   13    &   30    \\
\hline
C4.5      &   77.55       &   74.34     &  78.61  &   76.33 \\
3-nn      &   81.06       &   72.81     &  78.92  &   75.87 \\
RF        &   \bf{85.95}  &   85.33     &  85.94  &   \bf{86.71} \\
SVM       &   79.84       &   73.27     &  85.49  &   83.20 \\
\hline
mean       &   81.10      &   76.44     &  82.24  &   80.52 \\
\hline
\end{tabular}
\end{table}

In Tables \ref{resultsOriginal1PC} and \ref{resultsStandardized1PC} we present the accuracy for the algorithms using only one feature, that is, 1 principal component or the fractal dimension value. Also we show the results using 1 PC plus the fractal dimension value. From these Tables we can observe that random forest obtained five of the best results, while C4.5 obtained the other one. In addition we can see that, on average, classification using the fractal dimension value is better than using 1 principal component, considering standardized images. 

\begin{table}
\caption{Accuracy for original images using 1 principal component (1 PC) and the fractal dimension value (FDV). The best results are in bold.}
\label{resultsOriginal1PC}
\centering
\begin{tabular}{cccc} 
\hline
Algorithm  &   1 PC        &   FDV         & 1 PC + FDV  \\
\hline
C4.5       &  \bf{79.68}   &  77.09       &  \bf{79.38}   \\
3-nn       &  76.02        &  70.22       &  74.04        \\
RF         &  74.34        &  65.79       &  74.34        \\
SVM        &  79.38        &  \bf{79.38}  &  \bf{79.38}   \\
\hline
mean       &  77.35        &  73.12       &  76.78   \\
\hline
\end{tabular}
\end{table}

\begin{table}
\caption{Accuracy for standardized images using 1 principal component (1 PC) and the fractal dimension value (FDV). The best results are in bold.}
\label{resultsStandardized1PC}
\centering
\begin{tabular}{cccc} 
\hline
Algorithm  & 1 PC          &   FDV         &  1 PC + FDV  \\
\hline
C4.5       &   78.92       &  78.46       &   75.56   \\
3-nn       &   74.49       &  78.16       &   77.24   \\
RF         &   68.39       &  75.11       &   76.17   \\
SVM        &   \bf{79.38}  &  \bf{79.22}  &   \bf{78.62}   \\
\hline
mean       &   75.29  &  77.74 &  76.90   \\
\hline
\end{tabular}
\end{table}

\section{Discussion}
\label{sec:discussion}
\begin{figure}
\centering
\includegraphics[width=0.5\textwidth]{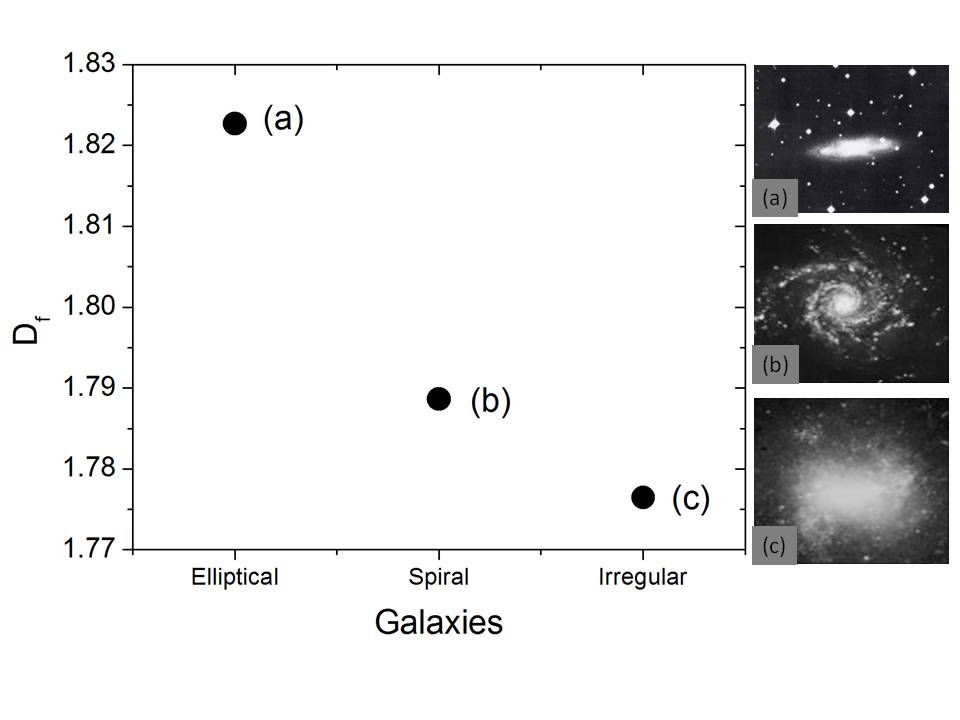}
\caption{\label{FDt} In (a) is shown an elliptical image which mean fractal dimension value is $D_{f}=1.8226$. In (b) is shown an image of an spiral galaxies, its fractal dimension mean value is $D_{f}=1.7886$; and finally in (c) is presented an irregular galaxy with a mean value of $D_{f}=1.7764$.}
\end{figure}

The difference of the fractal dimension values between the three types of galaxies analyzed is presented in Fig.~\ref{FDt}. The characterization by fractal dimension helps to distinguish with mathematical arguments the classification process of complicated images of galaxies. Results presented in Tables \ref{resultsOriginal} and \ref{resultsStandardized} show that the best results are obtained when standardized images and fractal dimension are used, particularly using random forest with 29 PCs plus the fractal dimension value. 
In addition we can observe that support vector machines was the algorithm with the greatest increase of accuracy when using PCs plus the fractal dimension value. That is from 73.27 to 85.29 per cent accuracy using 29 and 30 features, respectively. In fact,  in average across different classifiers, the performance improved my more than 4 per cent when including the fractal dimension as feature. 

In Tables \ref{confusionMatrixElliptical}, \ref{confusionMatrixSpiral} and \ref{confusionMatrixIrregular} we present the confusion matrix for the best algorithm to classify elliptical, spiral and irregular galaxies, respectively. From this results we can see that 3-nearest neighbors was the best algorithm to classify elliptical galaxies with 88.9 per cent accuracy; random forest was able to classify 100 per cent accuracy of the spiral galaxies; while C4.5 was the best algorithm to classify irregular galaxies with 40 per cent accuracy. We can also observe that none of the best results for elliptical and spiral galaxies have classified irregular galaxies correctly. On the other hand, when irregular galaxies are classified correctly, accuracy of elliptical decreases significantly; while the accuracy for spiral galaxies remains about 87 percent accuracy.

\begin{table}
\caption{Confusion matrix for the best algorithm to classify elliptical galaxies: 3-nearest neighbors.}
\label{confusionMatrixElliptical}
\centering
\begin{tabular}{lcccr} 
\hline
Galaxy type   &   Elliptical &  Spiral  & Irregular &  Accuracy per type\\
Elliptical    &       15     &      2   &    0      &    88.9 \%         \\
Spiral        &       12     &     92   &    0      &    88.4 \%         \\
Irregular     &        1     &      9   &    0      &       0 \%           \\
\hline
\end{tabular}
\end{table}

\begin{table}
\caption{Confusion matrix for the best algorithm to classify spiral galaxies: Random forest.}
\label{confusionMatrixSpiral}
\centering
\begin{tabular}{lcccr} 
\hline
Galaxy type   &   Elliptical &  Spiral  & Irregular &  Accuracy per type\\
Elliptical    &       11     &      6   &    0      &    64.7 \%         \\
Spiral        &       0      &    104   &    0      &   100.0 \%         \\
Irregular     &       0      &     10   &    0      &     0 \%           \\
\hline
\end{tabular}
\end{table}

\begin{table}
\caption{Confusion matrix for the best algorithm to classify irregular galaxies: C4.5.}
\label{confusionMatrixIrregular}
\centering
\begin{tabular}{lcccr} 
\hline
Galaxy type   &   Elliptical &  Spiral  & Irregular &  Accuracy per type\\
Elliptical    &       9      &      8   &    0      &    52.9 \%         \\
Spiral        &       8      &     91   &    5      &    87.5 \%         \\
Irregular     &       2      &      4   &    4      &    40.0 \%           \\
\hline
\end{tabular}
\end{table}

\section{Conclusions}
\label{sec:conclusions}
In this paper we have applied, for the first time, fractal dimension analysis to perform automatic morphological galaxy classification. The fractal dimension analysis contributes to distinguish the three main types of galaxies: elliptical, spiral and irregular. We found evidently differences between the rank of fractal dimension among the three groups of galaxies. By using the fractal dimension value as an additional attribute, it is possible to improve the classification accuracy, despite using a small set of images. The best results were obtained by 3-nearest neighbors and random forest using standardized images with PCs and the fractal dimension value. Future work includes: repeating the experiments using a larger data set of galaxy images; and creating ensemble of classifiers to identify each type of galaxy separately.

\section*{Acknowledgment}
EMCM thanks the financial support of CONACyT.

\bibliography{rsc} 
\bibliographystyle{rsc}

\end{document}